\documentclass[twoside,11pt]{article}

\usepackage{jmlr2e}
\usepackage{bbm}
\usepackage{amsmath}
\usepackage{amsfonts}
\usepackage{subcaption}
\usepackage{dsfont}

\renewcommand{\v}[1]{\boldsymbol{\mathbf{#1}}}
\DeclareMathOperator*{\argmax}{argmax}
\DeclareMathOperator*{\minimize}{minimize}
\DeclareMathOperator*{\subjectto}{subject\ to}
\newcommand{\E}{\mathbb{E}}
\newcommand{\var}{\mathbb{V}\mathrm{ar}}
\usepackage{diagbox}
\newcommand{\na}{\diagbox[dir=SW]{}{}}

\usepackage{lastpage}
\jmlrheading{27}{2026}{1-\pageref{LastPage}}{7/23; Revised 11/25}{2/26}{23-0958}{Jiangrong Ouyang, Mingming Gong, and Howard Bondell}

\ShortHeadings{Bayesian Inference of Contextual Bandit Policies}{Ouyang, Gong, and Bondell}
\firstpageno{1}

\begin{document}

\title{Bayesian Inference of Contextual Bandit Policies\\ via Empirical Likelihood}

\author{\name Jiangrong Ouyang \email jiangrong.ouyang@unimelb.edu.au \\
        \name Mingming Gong \email mingming.gong@unimelb.edu.au \\
        \name Howard Bondell \email howard.bondell@unimelb.edu.au \\
        \addr School of Mathematics and Statistics\\
                University of Melbourne\\
                Victoria, Australia}

\editor{Aurélien Garivier}

\maketitle

\begin{abstract}
Policy inference plays an essential role in the contextual bandit problem. In this paper, we use empirical likelihood to develop a Bayesian inference method for the joint analysis of multiple contextual bandit policies in finite sample regimes. The proposed inference method is robust to small sample sizes and is able to provide accurate uncertainty measurements for policy value evaluation. In addition, it allows for flexible inferences on policy comparison with full uncertainty quantification. We demonstrate the effectiveness of the proposed inference method using Monte Carlo simulations and its application to an adolescent body mass index data set.
\end{abstract}


\section{Introduction}
The contextual bandit problem is an extension of the standard multi-armed bandit problem, in which the rewards of arms depend on some context information \citep{auer_nonstochastic_2002, langford_epoch-greedy_2007}. A contextual bandit policy is a function that guides the player to select more profitable arms based on the given context information. The inference of contextual bandit policies is essential, since it can be used to improve the current policy or to experiment with hypothetical policies of interest. Hence, there is no need to actually run the hypothetical policies in the real world, since we are able to reconstruct the information just as what we would have observed if the desired policies had been applied.

To perform a reliable inference, one should first be able to evaluate policies accurately. However, the evaluation of a bandit policy is not straightforward, since only partial feedback is observed, and the available data is not necessarily collected using the policy we want to evaluate. Hence, we need to resort to off-policy evaluation \citep{sutton_reinforcement_2018}. Off-policy evaluation is a problem of evaluating a given target policy using the data collected with another policy, called behavior policy. 

There are three types of off-policy evaluation methods: direct method, importance sampling method, and doubly robust method. For the direct method, the expected reward given the arm and context is learned using regression, and then it is used to estimate the target policy directly. The direct method estimator relies heavily on the accuracy of the regression for the reward. If the regression model is misspecified, then the estimator can be biased and inconsistent. 

The main idea of the importance sampling method is to correct the weights of data points collected by the behavior policy so that they align with the target policy \citep{horvitz_generalization_1952}. For example, the importance sampling (IS) estimator takes the weighted average of the rewards, where the weights are given by the ratios of the probabilities from the target policy to that from the behavior policy. The IS estimator is unbiased and consistent, yet its variance is often large, especially when the target policy differs significantly from the behavior policy. The self-normalized IS (SNIS) estimator is a scaled IS estimator, normalized by the average of the importance weights, which decreases the mean squared error at the cost of being biased \citep{swaminathan_self-normalized_2015}. Furthermore, the SNIS estimator is bounded within the support of rewards, whereas the IS estimator is not.

The doubly robust estimator combines the direct method and importance sampling method, in which the direct method estimator acts as a baseline and is corrected by the IS estimator \citep{dudik_doubly_2011, jiang_doubly_2016}. If the regression model for the reward is specified correctly, then the mean squared error of the doubly robust estimator achieves the semiparametric lower bound asymptotically. Otherwise, it can be worse than that of the IS and SNIS estimators \citep{kang_demystifying_2007, kallus_intrinsically_2019}.

While many policy learning algorithms and evaluation methods are well-developed, policy inference remains difficult, since the distribution of the embedded importance weights is unknown. In this paper, we focus on the inference of contextual bandit policies. Empirical likelihood is a nonparametric inference method, which relies only on a set of estimating equations for the data \citep{owen_empirical_1988, owen_empirical_1990, qin_empirical_1994}. Hence, with appropriate estimating equations, inferences on the target policy can be drawn based on the empirical likelihood. It has been shown that the point estimator based on empirical likelihood generally has a lower mean squared error than the IS, SNIS, and DR estimators \citep{kallus_intrinsically_2019}. In addition, hypothesis tests and confidence intervals can be constructed using the empirical likelihood ratio with a chi-squared approximation \citep{karampatziakis_empirical_2020}. However, the empirical likelihood ratio in this setting only behaves like the chi-squared distribution when the sample size is large. For small or medium sample sizes, the confidence intervals are poorly calibrated and suffer from mismatched coverage probabilities. In addition, all the aforementioned methods focus on single policy evaluation, so comparison among potentially correlated policies is infeasible. 

Given the poor chi-squared calibration in the frequentist setting, we propose using empirical likelihood in the Bayesian paradigm to conduct proper inference, either with an uninformative prior or by accommodating any prior information about the policy value. Due to the fact that empirical likelihood is not an ordinary likelihood, the Bayesian inference with empirical likelihood does not follow automatically. \cite{lazar_bayesian_2003} explored the validity of the resultant posterior when the data density is replaced by empirical likelihood. \cite{grendar_asymptotic_2009} further justified the use of empirical likelihood in the Bayesian paradigm. Later, Bayesian empirical likelihood was adapted in several fields: survey sampling \citep{rao_bayesian_2010}, small area estimation \citep{chaudhuri_empirical_2011}, and quantile regression \citep{yang_bayesian_2012}. Algorithms for more efficient Bayesian computation have also been developed \citep{mengersen_bayesian_2013, chaudhuri_hamiltonian_2017, yu_variational_2024}.

In this paper, we employ empirical likelihood to develop a Bayesian inference method for jointly analyzing multiple contextual bandit policies in finite sample regimes. Our proposed inference method is robust with respect to small sample sizes and is able to provide accurate uncertainty measurements for policy value evaluation. Additionally, flexible inferences can be drawn on policy comparison with full uncertainty quantification. 

The remainder of this paper is organized as follows. Section~\ref{sec:background} provides the background information and defines the notation used throughout this paper. Section~\ref{sec:methodology} describes the proposed Bayesian empirical likelihood method for policy inference. Section~\ref{sec:experiments} presents two experiments demonstrating the proposed method. Section~\ref{sec:application} gives the application to an adolescent body mass index data set. Section~\ref{sec:discussion} contains some discussions. 

\section{Background and Notation}
\label{sec:background}
\textbf{Bandit Problem.} The multi-armed bandit problem is a repeated game proposed by \cite{robbins_aspects_1952}. In each round, the player makes a choice from $K$ different arms and receives the associated random reward. The goal is to develop a strategy to maximize the total rewards. This game is a paradigmatic example of the exploration–exploitation trade-off dilemma. As the player has no knowledge of the reward distribution associated with each arm initially, exploration is needed. Then the player can exploit the information acquired in the past to gain rewards at the moment or keep exploring for better rewards in the future. 

The contextual bandit problem is an extension of the standard multi-armed bandit problem. It further assumes that the reward distributions depend on some context information, and the context information is given to the player for decision making. This setting is more practical, since bandit problems in reality often come with context information. Due to the introduction of context information, contextual bandits have been adopted widely in various areas, such as news article recommendation \citep{li_contextual-bandit_2010}, personalized medicine \citep{tewari_ads_2017}, design of clinical trials \citep{villar_covariate-adjusted_2018}, and advertisement optimization \citep{narita_efficient_2019}. 

Consider the contextual bandit problem with $K$ arms. In each round, a context $\v x \in \mathcal{X}$ is generated from some distribution $D$. Given this context $\v x$, the reward vector $\v r = (r_1, r_2, \dots, r_K)^\top$ is drawn from the conditional distribution $D_{\v x}$. Based on the context $\v x$, the player selects an arm $a \in \{1, 2, \dots, K\}$, and receives the reward $r_a$. Note that only $r_a$ is revealed. Without loss of generality, we assume $r_a \in [0,1]$.

Let $\pi: \mathcal{X} \rightarrow \mathcal{P}$ be a contextual bandit policy, where $\mathcal{P}$ is the set of all probability mass functions over the arms. For a given context $\v x$, the policy $\pi$ guides the player to select an arm based on the probability mass function, $\pi_{\v x} = \pi(\v x)$; that is, $\pi_{\v x}(a)$ indicates the probability of choosing arm $a$ under the context $\v x$. The value of the policy $\pi$ is defined as the expected reward, i.e., 
\begin{equation*}
    \mathrm{v}_{\pi} = \E_{\v x \sim D, a \sim \pi_{\v x}} [r_a].
\end{equation*}

\textbf{Off-Policy Evaluation.} Consider the off-policy evaluation, where the offline data is collected with some behavior policy $p$ rather than the target policy $\pi$. That is, in each round, the arm is chosen according to the probability mass function $p_{\v x}$. We assume that $p_{\v x}$ is known or can be well estimated. 

Since the arms are chosen by the behavior policy $p$, for the associated rewards $r$, the target policy $\pi$ will have different probabilities of receiving them. To leverage these rewards, define the importance weight as the probability ratio, 
\begin{equation*}
    w_a = \frac{\pi_{\v x}(a)}{p_{\v x}(a)}.
\end{equation*}
It follows that the value of policy $\pi$ can be calculated by 
\begin{equation*}
    \mathrm{v}_{\pi} = \E_{\v x \sim D, a \sim p_{\v x}} [w_a r_a].
\end{equation*}
This is the main estimating equation we use to construct the empirical likelihood function. Note that the estimation and inference of policy $\pi$ depend on the offline data only through the importance weights and the received rewards, 
\begin{equation*}
    \{(w_i, r_i)\}_{i=1}^n,
\end{equation*}
where $w_i = \frac{\pi_{\v x_i}(a_i)}{p_{\v x_i}(a_i)}$ denotes the observed value of the importance weight in round $i$ in a data set of size $n$. 

\textbf{Empirical Likelihood.} Empirical likelihood is a nonparametric inference method proposed by \cite{owen_empirical_1988}. It does not require a specification of distributional assumptions for the data, but relies only on a set of estimating equations. Assume that $X_1, \dots, X_n \in \mathbb{R}$ are independent random variables with a common distribution function $F$. If the population mean, $\mu = \mu(F)$, is the parameter of interest, then one can use the first moment condition, $\E_F(X-\mu) = 0$, as the estimating equation, and define the log-empirical likelihood for $\mu$ as the solution to the following optimization problem, 
\begin{equation*}
    \log L(\mu) = 
    \sup_{\boldsymbol{\mathcal{Q}} \succeq \v 0} 
    \left\{ 
        \sum_{i=1}^n  \log(Q_i) \biggm| 
        \sum_{i=1}^n Q_i = 1, 
        \sum_{i=1}^n Q_i (X_i - \mu) = 0 
    \right\},
\end{equation*}
where $\boldsymbol{\mathcal{Q}} = (Q_1, Q_2, \dots, Q_n)^\top$.

More flexible data structures or restrictions on the parameters can be incorporated into the empirical likelihood function by adding extra estimating equations \citep{qin_empirical_1994}. Continue with the example above. If we know that the data is Poisson-like, then the second moment condition, $\E_F(X^2-\mu-\mu^2)=0$, can be included to ensure the property of $\var_F(X)=\E_F(X)$; that is, 
\begin{equation*}
    \log L(\mu) = 
    \sup_{\boldsymbol{\mathcal{Q}} \succeq \v 0} 
    \left\{ 
        \sum_{i=1}^n  \log(Q_i) \biggm| 
        \sum_{i=1}^n Q_i = 1, 
        \sum_{i=1}^n Q_i (X_i - \mu) = 0, 
        \sum_{i=1}^n Q_i \left(X_i^2 - \mu - \mu^2 \right) = 0 
    \right\}, 
\end{equation*}
where $\boldsymbol{\mathcal{Q}} = (Q_1, Q_2, \dots, Q_n)^\top$. 

Note that the probabilities are optimized and assigned only to the observed data points. For some parameter values, it is possible that none of the probability allocations satisfies the constraints specified in the optimization problem. In such cases, the log-empirical likelihoods are defined as $-\infty$.

\section{Methodology}
\label{sec:methodology}
In this section, we define the empirical likelihood for contextual bandit policies and demonstrate the proposed method for policy inference in the Bayesian paradigm. Additionally, we present a practical method for performing the computations required for the Bayesian inference.

\subsection{Joint Empirical Likelihood for Multiple Policies}
\label{sec:logLv}
Let $\v v \in [0,1]^{\ell}$ be the values of a set of $\ell$ policies, $\pi_1, \pi_2, \dots, \pi_{\ell}$. We define the empirical likelihood jointly for these values, $\v v$, with the estimating equation, 
\begin{equation*}
    \v v = \E_{\v x \sim D, a \sim p_{\v x}} [\v w_a r_a],
\end{equation*}
where 
\begin{equation*}
    \v w_a = 
    \left(
        \frac{\pi_{1, \v x}(a)}{p_{\v x}(a)}, 
        \frac{\pi_{2, \v x}(a)}{p_{\v x}(a)}, 
        \dots,
        \frac{\pi_{\ell, \v x}(a)}{p_{\v x}(a)}
    \right)^\top
\end{equation*}
is the importance weight vector for the $\ell$ policies, $\pi_1, \pi_2, \dots, \pi_{\ell}$, with respect to the behavior policy $p$. Alongside the estimating equation for $\v v$, we also include a constraint for the importance weight vector, 
\begin{equation*}
    \E_{\v x \sim D, a \sim p_{\v x}} [\v w_a] = \v 1. 
\end{equation*}
The reason for including this additional constraint is two-fold. While the theoretical expectation of $\v w$ is always equal to $\v 1$, the empirical version with respect to $(\v w, r)$ in the data space does not necessarily hold. By including this constraint in the formulation of the empirical likelihood, the average of $\v w$ over the data space is restricted to $\v 1$. As a consequence, the support of the empirical likelihood function is bounded and is equal to the true support, $\v v \in [0,1]^{\ell}$. In addition, this constraint guarantees that the empirical likelihood function is invariant to the metric used to measure the policies. Specifically, an equivalent way to assess policies is from the \textit{regret}, $\v u = \v 1 - \v v$, with the estimating equation, $\v u = \E_{\v x \sim D, a \sim p_{\v x}} [\v w_a ( 1-r_a)]$. The estimating equations for the \textit{value} and \textit{regret} yield equivalent results if and only if the constraint, $\E_{\v x \sim D, a \sim p_{\v x}} [\v w_a] = \v 1$, holds. 

The inclusion or non-inclusion of this constraint corresponds to the analogs of two well-known importance sampling approaches. If it is not included, then the maximum empirical likelihood estimator corresponds to the pure importance sampling estimator, which takes the weighted average of the rewards directly \citep{horvitz_generalization_1952}. If it is included, then the maximum empirical likelihood estimator corresponds to the self-normalized importance sampling estimator, where the empirical average of the importance weights is restricted to $1$ \citep{swaminathan_self-normalized_2015}. 

Unlike the conventional empirical likelihood, we consider not only distributions with mass at the data points, $\left\{(\v w_i, r_i)\right\}_{i=1}^n$, but all possible distributions defined over arbitrary points $(\v w, r)$ within the hyperrectangular support, 
\begin{equation*}
    \mathcal{S} = 
    \left(
        \prod_{j=1}^{\ell} [w_{j,\min}, w_{j,\max}]
    \right) 
    \times [0,1],
\end{equation*}
where $[w_{j,\min}, w_{j,\max}]$ is the range of importance weights inferred from policy $\pi_j$ and the behavior policy $p$. The use of all possible points in the support $\mathcal{S}$ ensures that the empirical likelihood is well-defined for all $\v v \in [0,1]^{\ell}$. 

Formally, we start from first principles and consider all probability measures $\v Q$ on $(\mathcal{S}, \mathcal{B}(\mathcal{S}))$. Let $Q_{\v w,r} = \v Q(\{(\v w, r)\})$. We define the log-empirical likelihood for the policy values as the solution to the following optimization problem,
\begin{equation}
    \log L(\v v) = 
    \sup_{\v Q} 
    \left\{ 
        \sum_{i=1}^n \log \left( Q_{\v w_i, r_i} \right)  \Biggm|
        \int_{\mathcal{S}} \v w \,\mathrm{d} \v Q(\v w,r) = \v 1, 
        \int_{\mathcal{S}} \v w r \,\mathrm{d} \v Q(\v w,r) = \v v 
    \right\}.
    \label{def:logLv}
\end{equation}

As a nonparametric maximum likelihood estimation problem, it suffices to consider only discrete $\v Q$ \citep{lindsay_mixture_1995}. Note that the optimization variable is the entire probability measure $\v Q$ on $(\mathcal{S}, \mathcal{B}(\mathcal{S}))$ rather than a probability vector over some fixed points. Hence, the set of the points $(\v w, r) \in \mathcal{S}$ that receive positive probability mass is not known \textit{a priori}, and identifying this set constitutes part of the optimization problem. 

By solving the Lagrange dual problem (see Appendix~\ref{sec:a1} for details), we get 
\begin{equation*}
    \log L (\v v) =  
    - \sup_{(\v \beta, \v \tau) \in \mathcal{V}} 
    \sum_{i=1}^n \log 
    \left(
        1 + \v \beta^\top (\v w_i - \v 1) + \v \tau^\top (\v w_i r_i - \v v)
    \right)
    + C,
\end{equation*}
where $C$ is an immaterial constant, $\v \beta$ and $\v \tau$ are the dual variables, and 
\begin{equation*}
    \mathcal{V} = 
    \left\{
        (\v \beta, \v \tau) \in \mathbb{R}^\ell \times \mathbb{R}^\ell \,\Big|\,  
        1 + \v \beta^\top (\v w - \v 1)+ \v \tau^\top (\v w r - \v v) \ge 0 \;\; 
        \forall (\v w, r) \in \mathcal{S} 
    \right\}.
\end{equation*}
While the inequality needs to hold for all $(\v w, r) \in \mathcal{S}$, it suffices to only consider the vertices of the hyperrectangular support $\mathcal{S}$, due to the linearity with respect to $\v \beta$ and $\v \tau$. 

As a consequence of considering all probability measures $\v Q$ on $(\mathcal{S}, \mathcal{B}(\mathcal{S}))$, the resulting empirical likelihood function may be flat over some region, implying that the maximum empirical likelihood estimator may not be unique (see Appendix~\ref{sec:a2} for details). However, this non-uniqueness does not affect the Bayesian inference that follows.

Using this joint empirical likelihood, we can derive the joint posterior distribution for $\v v$ in the Bayesian paradigm, and subsequently construct a credible region for $\v v$ from this distribution. Flexible inferences can be drawn with full uncertainty quantification. For example, we can compare a new policy against a baseline policy by computing $\mathbb{P}(\mathrm{v_{new} > v_{baseline}} + \delta)$ or $\mathbb{P}(\mathrm{v_{new}} > (1+\delta)\mathrm{v_{baseline}})$ for a desired margin $\delta$, or even $\mathbb{P}(g(\mathrm{v_{new}, v_{baseline}}) \ge 0)$ for an arbitrary function $g$. 

For inference on a single policy, the empirical likelihood defined above is equivalent to the dual likelihood proposed by \cite{karampatziakis_empirical_2020}. They used chi-squared approximation for the log-empirical likelihood ratio to construct confidence intervals for the policy value. In Section~\ref{sec:expt1}, we design an experiment to compare their confidence intervals in the frequentist framework to our credible intervals in the Bayesian paradigm.

\subsection{Empirical Likelihood for Policy Value Difference}
\label{sec:logLd}
Suppose we have a baseline policy and a new policy. We are interested in evaluating the potential improvement of the new policy compared to the baseline policy. Then, we can construct the empirical likelihood for the policy value difference, $d = \mathrm{v_{new} - v_{baseline}}$. Let $\v t = (-1, 1)^\top$ and $\v v = (\mathrm{v_{baseline}, v_{new}})^\top \in [0,1]^2$. We define the log-empirical likelihood for the difference as the solution to the following optimization problem,
\begin{equation}
    \log L(d) = 
    \sup_{\v Q} 
    \left\{ 
        \sum_{i=1}^n \log \left( Q_{\v w_i, r_i} \right)  \Biggm|
        \int_{\mathcal{S}} \v w \,\mathrm{d} \v Q(\v w,r) = \v 1, 
        \int_{\mathcal{S}} \v t^\top \v w r \,\mathrm{d} \v Q(\v w,r) = d 
    \right\}.
    \label{def:logLd}
\end{equation}

From the definitions \eqref{def:logLv} and \eqref{def:logLd}, it is clear that $\log L(d) = \sup_{\v v: \, d = \v t^\top \v v} \log L(\v v)$. Note that rather than work with the vector $\v v \in [0,1]^2$, we directly project $\v v$ onto a one-dimensional interval when formulating the empirical likelihood for $d$. This approach is more aligned with our interest and more computationally efficient. 

By solving the Lagrange dual problem (see Appendix~\ref{sec:a3} for details), we get
\begin{equation*}
    \log L(d) = 
    - \sup_{(\v \beta, \delta) \in \mathcal{D}} 
    \sum_{i=1}^n \log 
    \left(
        1 + \v \beta^\top (\v w_i - \v 1)+ \delta (\v t^\top \v w_i r_i - d)
    \right)
    + C,
\end{equation*}
where $C$ is an immaterial constant, $\v \beta$ and $\delta$ are the dual variables, and 
\begin{equation*}
    \mathcal{D} = 
    \left\{
        (\v \beta, \delta) \in \mathbb{R}^2 \times \mathbb{R} \,\Big|\,  
        1 + \v \beta^\top (\v w - \v 1)+ \delta (\v t^\top \v w r - d) \ge 0 \;\; 
        \forall (\v w, r) \in \mathcal{S} 
    \right\}.
\end{equation*}

Similar to the joint empirical likelihood for $\v v$, the inequality in the set $\mathcal{D}$ only needs to hold for the vertices of the hyperrectangular support $\mathcal{S}$. The maximum empirical likelihood estimator also may not be unique, though, as before, this does not affect the subsequent Bayesian inference.

With this empirical likelihood, we can derive the posterior distribution for the difference $d$ in the Bayesian paradigm. Policy comparison is now more straightforward, since $d$ is the quantity of interest. For example, we can determine $\mathbb{P}(\mathrm{v_{new} > v_{baseline}} + \delta)$ by computing $\mathbb{P}(d > \delta)$ from a one-dimensional posterior distribution. Note that this probability is not strictly identical to the one computed from the joint posterior distribution for $\v v$, since they are derived from different empirical likelihoods. However, they share similar values. See Section~\ref{sec:expt2} for an illustration. If the only interest is the policy value difference, we suggest using the univariate posterior distribution for $d$ to reduce the computational cost.

\subsection{Computational Method for Bayesian Inference}
\label{sec:sub-support}
For computing posterior probabilities, such as $\mathbb{P}(\mathrm{v_{new} > v_{baseline}} + \delta)$, we need to integrate the posterior distribution over a specific region. Here, we present a practical method for performing the computations required for the Bayesian inference. 

Since $\v v$ is bounded and typically in a low-dimensional space, we approximate the posterior on a fine grid, by evaluating the joint density and normalizing via summing over the grid. In high-dimensional cases, we recommend reducing the dimension of $\v v$ down to a small number of policy value comparisons. The evaluation process can be computationally intensive as each empirical likelihood value itself involves solving an optimization problem. Nevertheless, this can be done directly in parallel, which can avoid significant computational burden. 

However, as the sample size increases, the empirical likelihood concentrates on an arbitrarily small region, which renders any pre-defined grid not fine enough. Hence, the key is to define an adaptive sub-support for $\v v$ in which the empirical likelihood is larger than some small threshold. 

We define the sub-support for each dimension $\mathrm{v}_j$ as follows:
\begin{equation}
    \{\mathrm{v}_j \,|\, L(\v v) \ge c^{-1} L^* \},
    \label{def:sub-support}
\end{equation}
where $L^*$ is the maximum empirical likelihood over the full support, and $c$ is a large positive number. An adaptive sub-support for $\v v$ is formed by taking the Cartesian product of the sub-supports for all dimensions defined above. Although this sub-support may contain $\v v$ such that $L(\v v) < c^{-1} L^*$, it is the smallest hyperrectangular region in which $L(\v v) \ge c^{-1} L^*$. 

With this adaptive sub-support, we can define a fine grid, then compute posterior probabilities and credible intervals with high accuracy. Additionally, using such an adaptive sub-support avoids wasting time on the region in which the empirical likelihood is practically zero. For the policy value difference $d$, the same method is applied to determine its adaptive sub-support. 

See Appendix~\ref{sec:a4} for more details on the connection between the adaptive support and Wilks' confidence region (or interval), how we choose $c$, and the computation of the sub-support for each dimension $\mathrm{v}_j$ via an optimization problem. See also Section~\ref{sec:expt1} for a brief derivation of Wilks' confidence interval.

\section{Experiments}
\label{sec:experiments}

In this section, we design two experiments to demonstrate the proposed Bayesian empirical likelihood inference method. 

\textbf{Assumptions.} Consider a contextual bandit problem with $K=10$ arms. Let $\v x = (\v x_c, \v x_1, \dots, \v x_K)$ be the context information, where $\v x_c$ is a common context vector for all arms and $\v x_a$ is an arm-specific context vector for $a \in \{1, 2, \dots, K\}$. Assume that $\v x_c \in \mathbb{R}^d$ and $\v x_a \in \mathbb{R}^d$, where $d=12$ and 
\begin{align*}
    \v x_c &\sim \mathrm{Dirichlet}(1, \dots, 1) \\
    \v x_a &\sim \mathcal{N}(\v 0, I_d).
\end{align*}
Given a context $\v x$, the reward $r_a$ for each arm follows a Bernoulli distribution specified as follows:
\begin{align*}
    \eta_a &= \beta_0 + \beta_1 \v x_c^\top \v x_a \\
    p_a &= \frac{\exp(\eta_a)}{1+\exp(\eta_a)} \\
    r_a &\sim \mathrm{Ber}(p_a),
\end{align*}
where $\beta_0 = 0$ and $\beta_1=3$. Note that the rewards $r_a$ are correlated, since they all depend on the common context vector $\v x_c$. 

We use a completely randomized behavior policy to generate the offline data. That is, in each round, the arm is chosen uniformly at random, regardless of the context given. 

\textbf{Policy Learning.} We design a heuristic method for policy learning. For each arm, fit a logistic regression, using the reward $r_a$ as the response variable and $\v x_c^\top \v x_a$ as the covariate. Then estimate an upper bound $\mathrm{ub}_a$ for each arm as follows:
\begin{align*}
    \eta_a &= b_{0a} + b_{1a} \v x_c^\top \v x_a \\
    \mathrm{s.e.}(\eta_a) &= 
        \sqrt{
            [1, \v x_c^\top \v x_a] \, \mathcal{I}_a^{-1} [1, \v x_c^\top \v x_a]^\top
        }\\
    \mathrm{ub}_a &= 
        \frac
            {\exp(\eta_a + m \times \mathrm{s.e.}(\eta_a))}
            {1+\exp(\eta_a + m \times \mathrm{s.e.}(\eta_a))},
\end{align*}
where $(b_{0a}, b_{1a})$ are the coefficient estimates of the logistic regression, $\mathcal{I}_a$ is the Fisher information matrix, and $m \ge 0$ controls the confidence level for the upper bound estimate. 

A randomized policy $\pi$ is derived based on the collection of upper bound estimates for all arms. The probability of choosing arm $a$ is given by
\begin{equation*}
    \pi_{\v x}(a) = 
    \mathrm{ub}_a^s \bigg/
    \sum_{k=1}^K \mathrm{ub}_k^s,
\end{equation*}
where $s \ge 0$ is a hyperparameter of the probability mass function. If $s=0$, then $\pi$ is a completely randomized policy. For larger $s$, more probabilities are allocated on the arms with larger upper bound estimates. As $s \rightarrow \infty$, $\pi$ converges to a deterministic policy that chooses the arm with the highest upper bound estimate. We can further restrict the arms to be chosen to the best $K'$ arms by setting $\mathrm{ub}_a$ to zero if their values are ranked below $K'$. The three hyperparameters together, $(m, s, K')$, control the trade-off between exploration and exploitation. 

For inference purposes, we learn a baseline policy and a new policy by this heuristic method. For the baseline policy, we use a smaller sample size and set the hyperparameters for more exploration. In contrast, for the new policy, we use a larger sample size and set the hyperparameters for more exploitation. We compute the true policy values using Monte Carlo integration with one million random points. For reference, we also include the oracle policy that always chooses the best arm. See Table~\ref{table:policy_info} for a detailed summary. 

\begin{table}[ht]
\centering
\begin{tabular}{cccccc}
\hline\hline
Policy   & Size & $m$ & $s$ & $K'$ & True Value \\ \hline
Baseline & 256         & 1   & 2   & 3    & 0.75      \\
New      & 1024        & 1   & 1   & 1    & 0.83      \\
Oracle   & \na         & \na & \na & \na  & 0.84      \\ \hline
\end{tabular}
\caption{Policy learning information.}
\label{table:policy_info}
\end{table}

\textbf{Bayesian Inference.} Using the proposed Bayesian inference method, we first evaluate the baseline policy and the new policy separately, and then compare these two policies jointly. A flat prior is used throughout the experiments. All posterior computations are performed with a fine grid over the adaptive sub-support defined in Section~\ref{sec:sub-support}. We use a grid with $10^4$ points for one-dimensional posterior distributions, and a grid with $10^6$ points for the two-dimensional posterior distributions.

\subsection{Single Policy Inference}
\label{sec:expt1}

In the frequentist framework, confidence intervals can be used to quantify the uncertainty of a point estimate. \cite{karampatziakis_empirical_2020} constructed confidence intervals based on Wilks' theorem. Let $\mathrm{v}_0$ be the true policy value and $\hat{\mathrm{v}}$ be its maximum likelihood estimator. Then 
\begin{equation*}
    -2 \log 
    \left( 
        \frac{L(\mathrm{v}_0)}{L(\hat{\mathrm{v}})} 
    \right) 
    \stackrel{\mathrm{d}}{\rightarrow} \chi^2_1 \;\; 
    \mathrm{as} \; n \rightarrow \infty,
\end{equation*}
under mild regularity conditions. From this theorem, if $L(\mathrm{v})$ is much smaller than $L(\hat{\mathrm{v}})$, then the hypothesis that $\mathrm{v}_0 = \mathrm{v}$ is rejected. Equivalently, if $L(\mathrm{v})$ is larger than a certain threshold, then $\mathrm{v}$ is included in a confidence interval. Specifically, a $100(1-\alpha)\%$ Wilks' interval is defined as 
\begin{equation*}
    \left\{ 
        \mathrm{v} \,\Big|\, 
        L(\mathrm{v}) \ge 
        L(\hat{\mathrm{v}}) 
        \exp
        \left(
            - \,\frac{1}{2} F^{-1}_{\chi^2_1}(1-\alpha) 
        \right) 
    \right\},
\end{equation*}
where $F^{-1}_{\chi^2_1}(1-\alpha)$ denotes the $(1-\alpha)$-th quantile of the $\chi^2_1$ distribution.

A credible interval in the Bayesian paradigm is an analog of a confidence interval. Let $p_{\mathrm{post}}(\mathrm{v})$ be the posterior distribution of the policy value. A $100(1-\alpha)\%$ highest posterior density (HPD) interval, the narrowest type of credible interval, is defined as 
\begin{equation*}
    \{\mathrm{v} \,|\,  p_{\mathrm{post}}(\mathrm{v}) \ge c \},
\end{equation*}
where $c$ is the largest number such that 
\begin{equation*}
    \int_{
        \mathrm{v}: \, p_{\mathrm{post}}(\mathrm{v}) \ge c
    } 
    p_{\mathrm{post}}(\mathrm{v}) \,\mathrm{dv} = 
    1 - \alpha.
\end{equation*}

Note that the HPD and the Wilks' intervals share the same form, both being defined as an interval in which a certain threshold is exceeded. However, the HPD interval is determined by the posterior probability, while the Wilks' interval relies on a chi-squared approximation for the log-empirical likelihood ratio. 

A good interval estimate should have a coverage probability that closely matches the nominal level. An interval with an overcoverage probability may provide trivial information, often because it is too wide, while an interval with an undercoverage probability can provide misleading information. 

\begin{figure}[ht]
    \centering
    \begin{subcaptionblock}{\textwidth}
        \centering
        \includegraphics[width=\linewidth]{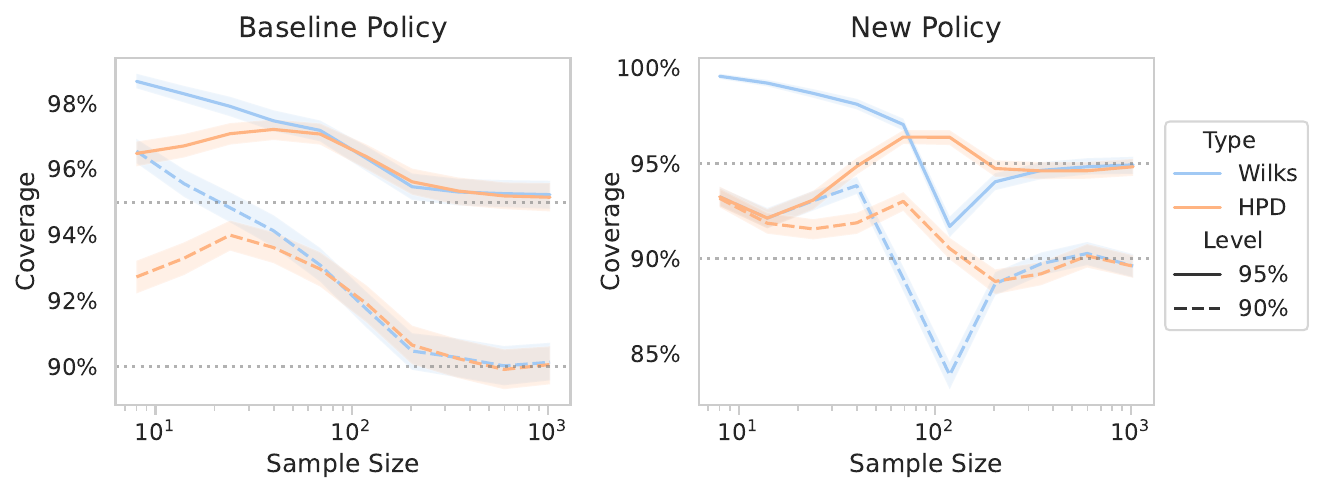}
        \vskip -0.08in
        \caption{Coverage probabilities for the Wilks' and the HPD intervals.}
        \label{fig:coverage}
    \end{subcaptionblock}
    \vskip 0.1in
    \begin{subcaptionblock}{\textwidth}
        \centering
        \includegraphics[width=\linewidth]{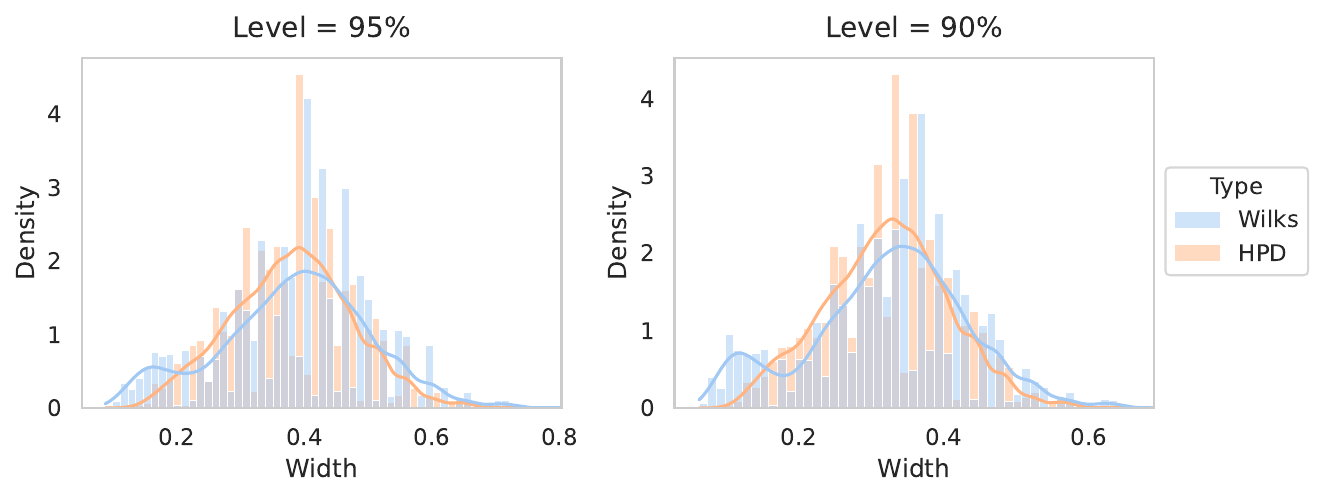}
        \vskip -0.08in
        \caption{Interval widths for the case where the Wilks' intervals suffer from undercoverage.}
        \label{fig:width}
    \end{subcaptionblock}%
    \caption{Comparison between the Wilks' and the HPD intervals.}
    \label{fig:coverage-width}
\end{figure}

In this experiment, we compute the Wilks' and the HPD intervals at the nominal levels of $90\%$ and $95\%$, and then compare their coverage probabilities and widths. Both the baseline policy and the new policy derived above are considered. Figure~\ref{fig:coverage} shows the coverage probabilities aggregated from $10,000$ replicates. As the sample size increases, the coverage probabilities of all intervals converge to the nominal levels. For the baseline policy, both the Wilks' and the HPD intervals exhibit overcoverage when the sample size is small, with the Wilks' intervals showing a stronger degree of overcoverage than the HPD intervals. For the new policy, it is evident that the coverage probabilities of the Wilks' intervals deviate significantly from the nominal levels when the sample size is small. 

In terms of the width, the Wilks' intervals are approximately $10\%$ wider than the HPD intervals on average when the sample size is small, but this margin quickly drops to zero as the sample size increases. Figure~\ref{fig:width} demonstrates the distributions of the interval widths for the case where the Wilks' intervals suffer from undercoverage. We observe that the widths of the Wilks' intervals spread over a larger range, with some intervals being quite wide and a significant proportion of intervals being too narrow. This explains the undercoverage problem. 

While both the Wilks' and the HPD intervals demonstrate good asymptotic properties, the HPD intervals are clearly more robust in finite sample regimes. The mismatched coverage probabilities and the variability in the widths of the Wilks' intervals can be explained by the poor chi-squared calibration when the sample size is small. For single policy inference, we recommend using the HPD intervals in the Bayesian paradigm for more accurate uncertainty quantification.

\subsection{Policy Comparison}
\label{sec:expt2} 

In this experiment, we compare the new policy against the baseline policy using both the joint posterior distribution for $\v v = (\mathrm{v_{baseline}, v_{new}})^\top$ and the univariate posterior distribution for $d = \mathrm{v_{new} - v_{baseline}}$. Specifically, we compute the probabilities of absolute improvement, $\mathbb{P}(\mathrm{v_{new} > v_{baseline}} + \delta)$, and the probabilities of relative improvement, $\mathbb{P}(\mathrm{v_{new}} > (1+\delta)\mathrm{v_{baseline}})$, for various margins $\delta = 0, 0.05, 0.10$. 

\begin{figure}[ht]
    \centering
    \includegraphics[width=\linewidth]{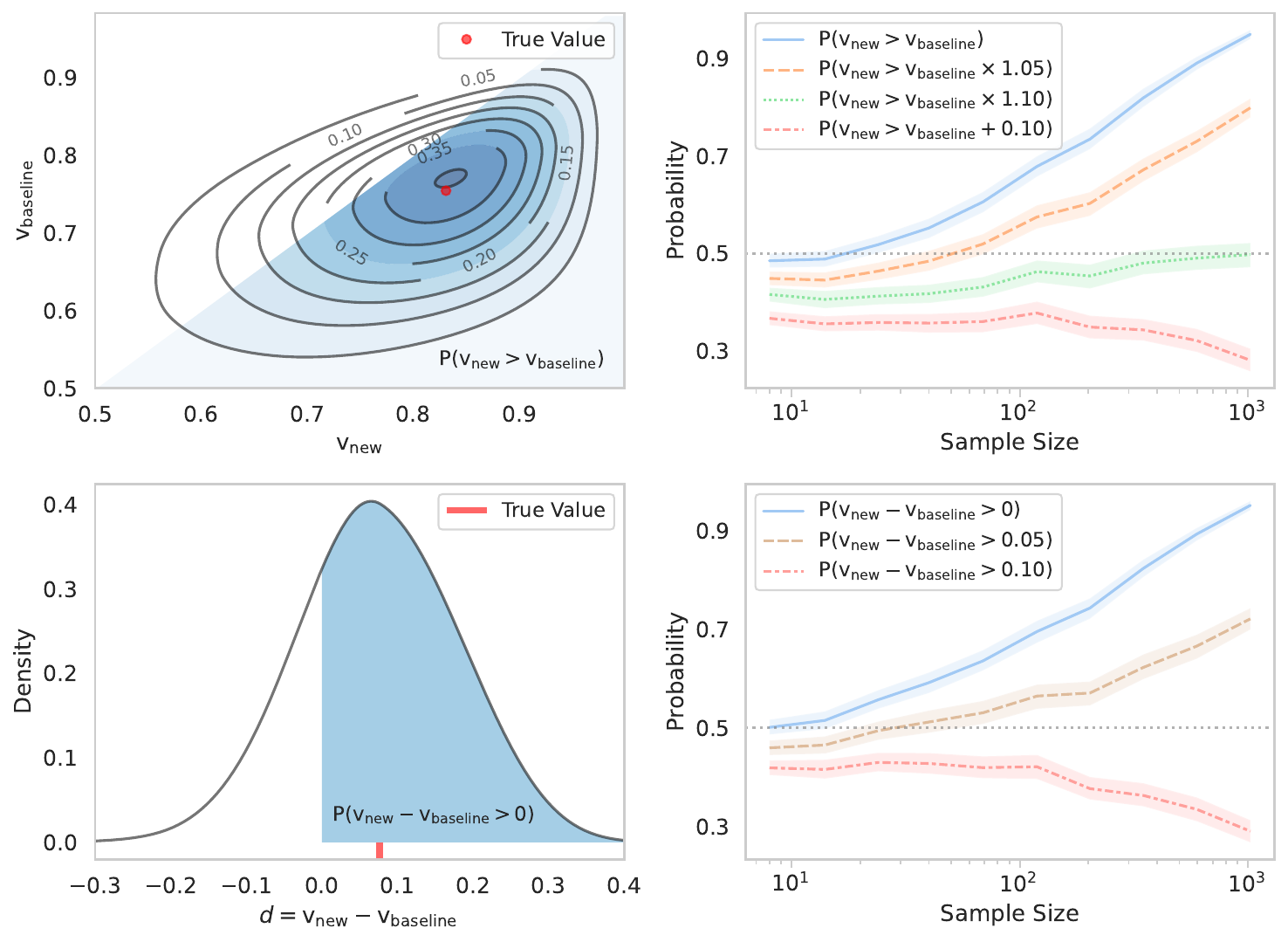}
    \caption{
        Policy comparison using the joint posterior distribution for 
            $\v v = (\mathrm{v_{baseline}, v_{new}})^\top$ (top) 
        and the univariate posterior distribution for 
            $d = \mathrm{v_{new} - v_{baseline}}$ (bottom).
    }
    \label{fig:comparison}
    \vskip -0.1in
\end{figure}

Figure~\ref{fig:comparison} demonstrates the comparison results using both methods. On the left, the colored regions illustrate the probabilities $\mathbb{P}(\mathrm{v_{new} > v_{baseline}})$ within the joint and univariate posterior distributions when the sample size is $100$. From the joint posterior distribution, it is clear that $\mathrm{v_{baseline}}$ and $\mathrm{v_{new}}$ are not independent, which suggests the necessity of the joint inference, either using the policy value vector $\v v$ or the policy value difference $d$. 

On the right of Figure~\ref{fig:comparison}, the curves show the probabilities of improvements for various margins, and the shaded areas represent the $95\%$ confidence bands based on the results aggregated from $500$ replicates. Since the joint posterior distribution for $\v v$ encompasses all information about the policy value vector, it supports arbitrary comparisons. We present the probabilities of both absolute and relative improvements; however, more flexible comparisons are possible. For example, we can compute the conditional probability $\mathbb{P}(\mathrm{v_{new} > v_{baseline} | v_{new}, v_{baseline} > 0.5})$. While the joint posterior distribution for $\v v$ allows for arbitrary comparisons, the univariate posterior distribution for $d$ provides a more direct method for comparing policies at lower computational cost. Depending on the interest, both methods offer effective policy comparisons with full quantification of uncertainty. 

\section{Application to Body Mass Index Data}
\label{sec:application}
Body mass index (BMI) is the ratio of human body weight (kg) to squared height ($\text{m}^2$). It was devised by Quetelet back to mid-19th century, and formalized by \cite{keys_indices_1972}. We demonstrate the proposed Bayesian empirical likelihood inference method with an adolescent BMI data set, generated by \cite{linn_iqlearn_2015} to mimic a two-stage randomized clinical trial of BMI reduction on adolescent obesity \citep{berkowitz_meal_2011}. Two treatments were used, meal replacement (MR) and conventional diet (CD). At both stages, patients were randomized to receive either MR or CD, each with probability 0.5. Four- and twelve-month patient BMI measurements were recorded. Four covariates are included: gender, race, parent BMI, and patient BMI measured at baseline. 

Consider a contextual bandit problem with two treatments, MR and CD. We use the data from the first stage only, since there is little change in the BMI measurements at the second stage. The data can be considered as being collected with a completely randomized policy, since patients were randomized to receive either treatment with equal probability. The median of the BMI reduction is 2.6, so we set the reward to 1 if the BMI reduces by more than 2.6, and 0 otherwise. 

A question of interest is what we would have observed if an alternative assignment of treatments had been applied to achieve more positive clinical outcomes. Our goal is to use the proposed method to address this question and guide subsequent decision-making with full quantification of uncertainty.

We set the current completely randomized policy as the baseline and define a new contextual bandit policy as follows. Partition data into two sets by the treatments chosen. For each treatment, fit a logistic regression with the four covariates. Then for each regression model, predict $\mathbb{P}(\mathrm{reward}=1)$ for all patients, and choose the treatment with the higher predicted estimate. Note that we expend no effort on tuning this heuristic policy, since our focus is on policy inference rather than on policy learning. 

We evaluate this new policy and compare it against the baseline policy using the proposed Bayesian empirical likelihood inference method. The point estimate of the new policy value is $0.64$, and the $95\%$ HPD interval is $(0.55, 0.72)$. From the joint posterior distribution for the policy value vector, $\mathbb{P}(\mathrm{v_{new} > v_{baseline}} \times 1.20) = 0.92$. From the univariate posterior distribution for the policy value difference, $\mathbb{P}(\mathrm{v_{new} > v_{baseline}} + 0.10) = 0.92$. Both metrics suggest a significant improvement in the policy value. These results demonstrate that the proposed method can effectively quantify uncertainty in policy comparisons and support subsequent decision-making in clinical treatment.

\section{Discussion}
\label{sec:discussion}

We propose a novel Bayesian empirical likelihood inference method for the joint analysis of multiple contextual bandit policies. While most existing techniques focus on policy learning and optimization, formal and effective inference on the learned policies is also necessary to complement these approaches. Our inference method provides accurate uncertainty measurements for policy value evaluation and facilitates flexible policy comparisons with full uncertainty quantification. Although it does not directly perform policy optimization, the effective inference results can be used to choose among multiple candidate policies learned, and consequently aid in the search for an optimal policy. 

By constructing the joint empirical likelihood for the policy value vector, arbitrary comparisons among potentially correlated policies are made possible. In addition, we construct the empirical likelihood for the policy value difference, which enables a more straightforward comparison at lower computational cost. 

The method for the policy value difference can be further generalized to accommodate other policy comparisons. For example, if $\mathrm{(v_3 - v_1, v_3 - v_2)^\top}$ or $\mathrm{v_3 - (v_1 + v_2)/2}$ is of interest, then we can apply the corresponding transformation on $\v v$ and construct the empirical likelihood in a lower-dimensional space. 

For computational efficiency, we develop an adaptive grid method for the posterior computations. However, it may still be inefficient in high dimensions. One resolve for this issue is to transform $\v v$ to a lower-dimensional space as discussed above. Alternatively, one can resort to variational inference \citep{blei_variational_2017, yu_variational_2024}. However, a loss of accuracy arising from the variational approximation is inevitable. Further investigation is required to control the loss of accuracy.


\acks{This research was partially funded by the Australian Government through the Australian Research Council.}



\appendix
\section{Joint Empirical Likelihood for Multiple Policies}
\label{sec:a1}

Let $\v v \in [0,1]^{\ell}$ be the values of a set of $\ell$ policies, $\pi_1, \pi_2, \dots, \pi_{\ell}$. As in \eqref{def:logLv} of Section~\ref{sec:logLv}, we define $-\log L (\v v)$ as the solution to the following optimization problem,
\begin{equation*}
    - \log L (\v v) = 
    \inf_{\v Q} 
    \left\{ 
        - \sum_{i=1}^n \log \left( Q_{\v w_i, r_i} \right) \Biggm| 
        \int_{\mathcal{S}} \v w \,\mathrm{d} \v Q(\v w,r) = \v 1, 
        \int_{\mathcal{S}} \v w r \,\mathrm{d} \v Q(\v w,r) = \v v 
    \right\},
\end{equation*}
where $\mathcal{S} = \left(\prod_{j=1}^{\ell} [w_{j,\min}, w_{j,\max}]\right) \times [0,1]$, $\v Q$ is a probability measure on $(\mathcal{S}, \mathcal{B}(\mathcal{S}))$, and $Q_{\v w_i,r_i} = \v Q(\{(\v w_i, r_i)\})$. 

As a nonparametric maximum likelihood estimation problem, it suffices to consider only discrete probability measures $\v Q$ \citep{lindsay_mixture_1995}. Let $\mathcal{A}$ denote the set of points in $\mathcal{S}$ with positive probability mass, 
\begin{equation*}
    \mathcal{A} = \{ (\v w,r) \in \mathcal{S}  \,|\, \v Q(\{(\v w, r)\}) > 0 \}. 
\end{equation*}
Then $-\log L (\v v)$ can be rewritten as
\begin{alignat*}{2}
    & \minimize_{\v Q} \quad && - \sum_{(\v w, r) \in \mathcal{A}} c_{\v w, r} \log \left( Q_{\v w, r} \right) \\
    & \subjectto             &&   \sum_{(\v w, r) \in \mathcal{A}} Q_{\v w,r} = 1 \\
    &                        &&   \sum_{(\v w, r) \in \mathcal{A}} \v w Q_{\v w,r} = \v 1 \\
    &                        &&   \sum_{(\v w, r) \in \mathcal{A}} \v w r Q_{\v w,r} = \v v,
\end{alignat*}
where $c_{\v w,r} = \sum_{i=1}^n \mathds{1}_{ \{ \v w=\v w_i, r=r_i \} }$. 

The associated Lagrange dual function is given by
\begin{align*}
    g(\gamma, \v \beta, \v \tau)
    &= \begin{aligned}[t]
    \inf_{\v Q} 
        \left\{
            - \sum_{(\v w, r) \in \mathcal{A}} c_{\v w, r} \log \left( Q_{\v w, r} \right) \right.
            &+ \gamma 
            \left( 
                \sum_{(\v w, r) \in \mathcal{A}} Q_{\v w,r} - 1 
            \right) \\
            &+ \v \beta^\top 
            \left( 
                \sum_{(\v w, r) \in \mathcal{A}} \v w Q_{\v w,r} - \v 1 
            \right) \\
            &+ \left.
            \v \tau^\top 
            \left( 
                \sum_{(\v w, r) \in \mathcal{A}} \v w r Q_{\v w,r} - \v v 
            \right) 
        \right\}
    \end{aligned} \\
    &= \inf_{\v Q} 
    \left\{ 
        \vphantom{ \sum_{(\v w, r) \in \mathcal{A}} }
        - \left( \gamma + \v \beta^\top \v 1 + \v \tau^\top \v v \right) \right. \\
    &\hspace{3cm}
        \left. + \sum_{(\v w, r) \in \mathcal{A}} 
        \left\{ 
            \left( 
                \gamma + \v \beta^\top \v w + \v \tau^\top \v w r 
            \right) Q_{\v w, r} 
            - c_{\v w, r} \log \left( Q_{\v w, r} \right) 
        \right\} 
    \right\}. 
\end{align*}

To ensure that the Lagrange dual function is bounded, we must have 
\begin{align*}
    & \gamma + \v \beta^\top \v w + \v \tau^\top \v w r \ge 0 \quad 
        \forall (\v w, r) \in \mathcal{A} \subset \mathcal{S}\\
    & \gamma + \v \beta^\top \v w + \v \tau^\top \v w r > 0 \quad 
        \mathrm{if} \; c_{\v w,r} > 0.
\end{align*}
Note that $\mathcal{A}$ is an arbitrary subset of $\mathcal{S}$. Since we require boundedness for any such $\mathcal{A} \subset \mathcal{S}$, the constraints above must therefore hold for all $(\v w, r) \in \mathcal{S}$. Now, if $\gamma + \v \beta^\top \v w + \v \tau^\top \v w r = 0$, then $c_{\v w,r} = 0$, and the corresponding summand involving $Q_{\v w,r}$ vanishes from the Lagrange dual function. Hence, the only relevant allocation of $Q_{\v w, r}$ with respect to the infimum is when $\gamma + \v \beta^\top \v w + \v \tau^\top \v w r > 0$. Using standard calculus for minimization, we get
\begin{equation*}
    Q^*_{\v w,r} = \frac{c_{\v w, r}}{\gamma + \v \beta^\top \v w + \v \tau^\top \v w r}.
\end{equation*}

Hence, the Lagrange dual function becomes
\begin{align*}
    g(\gamma, \v \beta, \v \tau)
    &= 
    - \left( \gamma + \v \beta^\top \v 1 + \v \tau^\top \v v \right)
    + \sum_{(\v w, r) \in \mathcal{A}} 
        \left\{ 
            c_{\v w, r} - c_{\v w, r} \log 
            \left( 
                \frac{c_{\v w, r}}{\gamma + \v \beta^\top \v w + \v \tau^\top \v w r} 
            \right)
        \right\} \\
    &= 
    - \left( \gamma + \v \beta^\top \v 1 + \v \tau^\top \v v \right)
    + n 
    + \sum_{(\v w, r) \in \mathcal{A}} 
        \left\{ 
            - c_{\v w, r} \log 
            \left( 
                \frac{c_{\v w, r}}{\gamma + \v \beta^\top \v w + \v \tau^\top \v w r} 
            \right)
        \right\}. 
\end{align*}

Now consider
\begin{equation*}
    \sum_{(\v w, r) \in \mathcal{A}} 
    (\gamma + \v \beta^\top \v w + \v \tau^\top \v w r) Q^*_{\v w,r}. 
\end{equation*}
From the $Q^*_{\v w,r}$ obtained above, this sum equals 
\begin{equation*}
    \sum_{(\v w, r) \in \mathcal{A}} c_{\v w, r} = n.
\end{equation*}
But, expanding the same sum and then applying the primal constraints yields
\begin{equation*}
    \gamma \sum_{(\v w, r) \in \mathcal{A}} Q^*_{\v w,r} 
        + \v \beta^\top \sum_{(\v w, r) \in \mathcal{A}} \v w Q^*_{\v w,r} 
        + \v \tau^\top \sum_{(\v w, r) \in \mathcal{A}} \v w r Q^*_{\v w,r} 
    = \gamma + \v \beta^\top \v 1 + \v \tau^\top \v v,
\end{equation*}
where the equality holds when the dual variables are evaluated at their optimal values. Hence, $\gamma + \v \beta^\top \v 1 + \v \tau^\top \v v = n$. This identity allows us to eliminate $\gamma$. We reparameterize the Lagrange dual function as 
\begin{equation*}
    \tilde{g}(\v \beta, \v \tau) = 
    g(n - \v \beta^\top \v 1 - \v \tau^\top \v v, \; \v \beta, \; \v \tau), 
\end{equation*}
which simplifies to
\begin{equation*}
    \tilde{g}(\v \beta, \v \tau) 
    = \sum_{(\v w, r) \in \mathcal{A}} 
    c_{\v w, r} \log 
    \left(
        n + \v \beta^\top (\v w - \v 1)+ \v \tau^\top (\v w r - \v v)
    \right) 
    - \sum_{(\v w, r) \in \mathcal{A}} c_{\v w, r} \log(c_{\v w, r}).
\end{equation*}

Let $\v \beta \leftarrow n \v \beta$ and $\v \tau \leftarrow n \v \tau$, rescaling the dual variables to maintain numerical stability as the sample size $n$ increases. Then by strong duality, $-\log L (\v v)$ can be obtained by solving the Lagrange dual problem,
\begin{align*}
    - \log L (\v v) 
    &= \sup_{(\v \beta, \v \tau) \in \mathcal{V}} \tilde{g}(\v \beta, \v \tau) \\
    &= \sup_{(\v \beta, \v \tau) \in \mathcal{V}} \sum_{(\v w, r) \in \mathcal{A}} 
        c_{\v w, r} \log 
        \left(
            1 + \v \beta^\top (\v w - \v 1) + \v \tau^\top (\v w r -  \v v)
        \right)
        - C \\
    &= \sup_{(\v \beta, \v \tau) \in \mathcal{V}} \sum_{i=1}^n 
        \log 
        \left(
            1 + \v \beta^\top (\v w_i - \v 1) + \v \tau^\top (\v w_i r_i -  \v v)
        \right)
        - C,
\end{align*}
where $C = \sum_{(\v w, r) \in \mathcal{A}} c_{\v w, r} \log(c_{\v w, r}) - n \log(n)$ is an immaterial constant, and 
\begin{equation*}
    \mathcal{V} = 
    \left\{
        (\v \beta, \v \tau) \in \mathbb{R}^\ell \times \mathbb{R}^\ell  
        \,\Big|\, 
        1 + \v \beta^\top (\v w - \v 1) + \v \tau^\top (\v w r - \v v) \ge 0 
        \;\; \forall (\v w, r) \in \mathcal{S} 
    \right\}.
\end{equation*}


\section{Maximum Empirical Likelihood Estimator}
\label{sec:a2}

Let $\hat{\v v}$ be the maximum empirical likelihood estimator, i.e.,
\begin{equation*}
    \hat{\v v} = \argmax_{\v v} \log L(\v v).
\end{equation*}
Recall that $\log L(\v v)$ itself is defined as the maximum of an optimization problem, where the parameter $\v v$ is incorporated through the estimating equation, $\v v = \int_{\mathcal{S}} \v w r \,\mathrm{d} \v Q(\v w, r)$, and the probability measure $\v Q$ is the optimization variable. For a fixed value $\v v$, this equation results in a constraint on the set of feasible $\v Q$ to be considered in the optimization. This is what creates the likelihood surface as $\v v$ varies. To compute the maximum empirical likelihood estimator, the problem becomes optimizing over all possible values of this surface, which occurs when the set of feasible $\v Q$ is unconstrained with respect to $\v v$. Then the maximum empirical likelihood estimator can be obtained by plugging the overall optimal $\v Q^*$ into the estimating equation, i.e., $\hat{\v v}  = \int_{\mathcal{S}} \v w r \,\mathrm{d} \v Q^*(\v w, r)$. 

Consider the optimization problem for the negative of the maximum log-empirical likelihood, 
\begin{equation*}
    - \log L^* = 
    \inf_{\v Q} 
    \left\{ 
        - \sum_{i=1}^n \log \left( Q_{\v w_i, r_i} \right)  \Biggm|
        \int_{\mathcal{S}} \v w \,\mathrm{d} \v Q(\v w,r) = \v 1 
    \right\},
\end{equation*}
where $\mathcal{S} = \left(\prod_{j=1}^{\ell} [w_{j,\min}, w_{j,\max}]\right) \times [0,1]$, $\v Q$ is a probability measure on $(\mathcal{S}, \mathcal{B}(\mathcal{S}))$, and $Q_{\v w_i,r_i} = \v Q(\{(\v w_i, r_i)\})$. 

As a nonparametric maximum likelihood estimation problem, it suffices to consider only discrete probability measures $\v Q$ \citep{lindsay_mixture_1995}. Let $\mathcal{A}$ denote the set of points in $\mathcal{S}$ with positive probability mass, 
\begin{equation*}
    \mathcal{A} = \{ (\v w,r) \in \mathcal{S}  \,|\, \v Q(\{(\v w, r)\}) > 0 \}. 
\end{equation*}
Then $-\log L^*$ can be rewritten as
\begin{alignat*}{2}
    & \minimize_{\v Q} \quad && - \sum_{(\v w, r) \in \mathcal{A}} c_{\v w, r} \log \left( Q_{\v w, r} \right) \\
    & \subjectto             &&   \sum_{(\v w, r) \in \mathcal{A}} Q_{\v w,r} = 1 \\
    &                        &&   \sum_{(\v w, r) \in \mathcal{A}} \v w Q_{\v w,r} = \v 1,
\end{alignat*}
where $c_{\v w,r} = \sum_{i=1}^n \mathds{1}_{ \{ \v w=\v w_i, r=r_i \} }$. 

The associated Lagrange dual function is given by
\begin{align*}
    g(\gamma, \v \beta)
    &= \begin{aligned}[t]
    \inf_{\v Q} 
        \left\{
            - \sum_{(\v w, r) \in \mathcal{A}} c_{\v w, r} \log \left( Q_{\v w, r} \right) \right.
            &+ \gamma 
            \left( 
                \sum_{(\v w, r) \in \mathcal{A}} Q_{\v w,r} - 1 
            \right) \\
            &+ \left.
            \v \beta^\top 
            \left( 
                \sum_{(\v w, r) \in \mathcal{A}} \v w Q_{\v w,r} - \v 1 
            \right)
        \right\}
    \end{aligned} \\
    &= \inf_{\v Q} 
    \left\{ 
        - \left( \gamma + \v \beta^\top \v 1 \right)
        + \sum_{(\v w, r) \in \mathcal{A}} 
        \left\{ 
            (\gamma + \v \beta^\top \v w) Q_{\v w, r} 
            - c_{\v w, r} \log \left( Q_{\v w, r} \right) 
        \right\} 
    \right\}.
\end{align*}

To ensure that the Lagrange dual function is bounded, we must have 
\begin{align*}
    & \gamma + \v \beta^\top \v w \ge 0 \quad 
        \forall (\v w, r) \in \mathcal{A} \subset \mathcal{S}\\
    & \gamma + \v \beta^\top \v w > 0 \quad 
        \mathrm{if} \; c_{\v w,r} > 0.
\end{align*}
Note that $\mathcal{A}$ is an arbitrary subset of $\mathcal{S}$. Since we require boundedness for any such $\mathcal{A} \subset \mathcal{S}$, the constraints above must therefore hold for all $(\v w, r) \in \mathcal{S}$. Now, if $\gamma + \v \beta^\top \v w = 0$, then $c_{\v w,r} = 0$, and the corresponding summand involving $Q_{\v w,r}$ vanishes from the Lagrange dual function. Hence, the only relevant allocation of $Q_{\v w, r}$ with respect to the infimum is when $\gamma + \v \beta^\top \v w > 0$. Using standard calculus for minimization, we get
\begin{equation*}
    Q^*_{\v w,r} = \frac{c_{\v w, r}}{\gamma + \v \beta^\top \v w}.
\end{equation*}

Hence, the Lagrange dual function becomes
\begin{align*}
    g(\gamma, \v \beta)
    &= 
    - \left( \gamma + \v \beta^\top \v 1 \right)
    + \sum_{(\v w, r) \in \mathcal{A}} 
        \left\{ 
            c_{\v w, r} - c_{\v w, r} \log 
            \left( 
                \frac{c_{\v w, r}}{\gamma + \v \beta^\top \v w } 
            \right)
        \right\} \\
    &= 
    - \left( \gamma + \v \beta^\top \v 1 \right)
    + n 
    + \sum_{(\v w, r) \in \mathcal{A}} 
        \left\{ 
            - c_{\v w, r} \log 
            \left( 
                \frac{c_{\v w, r}}{\gamma + \v \beta^\top \v w } 
            \right)
        \right\}. 
\end{align*}

Now consider
\begin{equation*}
    \sum_{(\v w, r) \in \mathcal{A}} 
    (\gamma + \v \beta^\top \v w) Q^*_{\v w,r}. 
\end{equation*}
From the $Q^*_{\v w,r}$ obtained above, this sum equals 
\begin{equation*}
    \sum_{(\v w, r) \in \mathcal{A}} c_{\v w, r} = n.
\end{equation*}
But, expanding the same sum and then applying the primal constraints yields
\begin{equation*}
    \gamma \sum_{(\v w, r) \in \mathcal{A}} Q^*_{\v w,r} 
        + \v \beta^\top \sum_{(\v w, r) \in \mathcal{A}} \v w Q^*_{\v w,r} 
    = \gamma + \v \beta^\top \v 1, 
\end{equation*}
where the equality holds when the dual variables are evaluated at their optimal values. Hence, $\gamma + \v \beta^\top \v 1 = n$. This identity allows us to eliminate $\gamma$. We reparameterize the Lagrange dual function as 
\begin{equation*}
    \tilde{g}(\v \beta) = 
    g(n - \v \beta^\top \v 1, \; \v \beta), 
\end{equation*}
which simplifies to
\begin{equation*}
    \tilde{g}(\v \beta) 
    = \sum_{(\v w, r) \in \mathcal{A}} 
    c_{\v w, r} \log 
    \left(
        n + \v \beta^\top (\v w - \v 1)
    \right) 
    - \sum_{(\v w, r) \in \mathcal{A}} c_{\v w, r} \log(c_{\v w, r}).
\end{equation*}

Let $\v \beta \leftarrow n \v \beta$, rescaling the dual variables to maintain numerical stability as the sample size $n$ increases. Then by strong duality, $-\log L^*$ can be obtained by solving the Lagrange dual problem,
\begin{align*}
    - \log L^* 
    &= \sup_{\v \beta \in \mathcal{V}'} \tilde{g}(\v \beta) \\
    &= \sup_{\v \beta \in \mathcal{V}'} \sum_{(\v w, r) \in \mathcal{A}} 
        c_{\v w, r} \log 
        \left(
            1 + \v \beta^\top (\v w - \v 1) 
        \right) 
        - C \\
    &= \sup_{\v \beta \in \mathcal{V}'} \sum_{i=1}^n 
        \log 
        \left(
            1 + \v \beta^\top (\v w_i - \v 1) 
        \right) 
        - C,
\end{align*}
where $C = \sum_{(\v w, r) \in \mathcal{A}} c_{\v w, r} \log(c_{\v w, r}) - n \log(n)$ is an immaterial constant, and 
\begin{equation*}
    \mathcal{V}' = 
    \left\{
        \v \beta \in \mathbb{R}^{\ell}  
        \,\Big|\, 
        1 + \v \beta^\top (\v w - \v 1) \ge 0 
        \;\; \forall (\v w, r) \in \mathcal{S} 
    \right\}.
\end{equation*}

After solving this dual problem, we can use the optimal dual variable, $\v \beta^*$, to construct the maximum empirical likelihood estimator, $\hat{\v v}$. Recall that the policy value vector depends on the distribution of the mass over the hyperrectangular support, i.e., 
\begin{equation*}
    \v v 
    = \int_{\mathcal{S}} \v w r \,\mathrm{d} \v Q(\v w, r) 
    = \sum_{(\v w, r) \in \mathcal{A}} \v w r Q_{\v w, r}.
\end{equation*}
The probabilities assigned to the observed data points are given by 
\begin{equation*}
    Q^*_{\v w_i,r_i} = \frac{1}{n(1 + \v \beta^{*\top} (\v w_i- \v 1))}.
\end{equation*}
The remaining probabilities, if any, are allocated to the boundary points of $\mathcal{S}$ that satisfies 
\begin{equation*}
    1 + \v \beta^{*\top} (\v w - \v 1) = 0. 
\end{equation*}
Note that this condition is free of the reward, $r$. Hence, for any $\v w_0$ that satisfies the condition, probabilities can be assigned to points over the interval $\{(\v w_0, r) \, | \, r \in [0,1]\}$. This explains why the maximum empirical likelihood estimator may not be unique.

Let 
\begin{equation*}
    \mathcal{W}_0 = 
    \left\{ 
        \v w_0 \in \prod_{j=1}^{\ell} [w_{j,\min}, w_{j,\max}] \Biggm| 
        1 + \v \beta^{*\top} (\v w_0 - \v 1) = 0
    \right\}.
\end{equation*}
Then
\begin{equation*}
    \hat{\v v} = 
    \left\{ 
        \sum_{i=1}^n \v w_i r_i Q^*_{\v w_i,r_i} 
        + \sum_{\v w_0 \in \mathcal{W}_0} \v w_0 r_{\v w_0} Q^*_{\v w_0} \Biggm| 
        \forall r_{\v w_0} \in [0, 1] 
    \right\},
\end{equation*}
where $Q^*_{\v w_0}$ are solved from the primal conditions, 
\begin{align*}
    &\sum_{\v w_0 \in \mathcal{W}_0} Q_{\v w_0} 
        = 1 - \sum_{i=1}^n Q^*_{\v w_i,r_i}\\
    &\sum_{\v w_0 \in \mathcal{W}_0} \v w_0 Q_{\v w_0} 
        = \v 1 - \sum_{i=1}^n \v w_i Q^*_{\v w_i,r_i}.
\end{align*}

Note that $\mathcal{W}_0$ may be empty, which occurs if $\sum_{i=1}^n Q^*_{\v w_i,r_i} = 1$. In this case, $\hat{\v v}$ is unique. If $\mathcal{W}_0$ is not empty, then due to the linearity with respect to $\v \beta$, it suffices to consider those elements of $\mathcal{W}_0$ that are the vertices of $\prod_{j=1}^{\ell} [w_{j,\min}, w_{j,\max}]$ when constructing $\hat{\v v}$. For the same reason, for any candidate $\v w_0 \in \mathcal{W}_0$, it suffices to consider $(\v w_0, 0)$ and $(\v w_0, 1)$.

\section{Empirical Likelihood for Policy Value Difference}
\label{sec:a3}

Let $\v t = (-1, 1)^\top$, $\v v = (\mathrm{v_{baseline}, v_{new}})^\top \in [0,1]^2$, and $d = \v t^\top \v v$. As in \eqref{def:logLd} of Section~\ref{sec:logLd}, we define $-\log L(d)$ as the solution to the following optimization problem, 
\begin{equation*}
    - \log L(d) = 
    \inf_{\v Q} 
    \left\{ 
        - \sum_{i=1}^n \log \left( Q_{\v w_i, r_i} \right)  \Biggm|
        \int_{\mathcal{S}} \v w \,\mathrm{d} \v Q(\v w,r) = \v 1, 
        \int_{\mathcal{S}} \v t^\top \v w r \,\mathrm{d} \v Q(\v w,r) = d 
    \right\}, 
\end{equation*}
where $\mathcal{S} = \left(\prod_{j=1}^{\ell} [w_{j,\min}, w_{j,\max}]\right) \times [0,1]$, $\v Q$ is a probability measure on $(\mathcal{S}, \mathcal{B}(\mathcal{S}))$, and $Q_{\v w_i,r_i} = \v Q(\{(\v w_i, r_i)\})$. 

As a nonparametric maximum likelihood estimation problem, it suffices to consider only discrete probability measures $\v Q$ \citep{lindsay_mixture_1995}. Let $\mathcal{A}$ denote the set of points in $\mathcal{S}$ with positive probability mass, 
\begin{equation*}
    \mathcal{A} = \{ (\v w,r) \in \mathcal{S}  \,|\, \v Q(\{(\v w, r)\}) > 0 \}. 
\end{equation*}
Then $-\log L (d)$ can be rewritten as
\begin{alignat*}{2}
    & \minimize_{\v Q} \quad && - \sum_{(\v w, r) \in \mathcal{A}} c_{\v w, r} \log \left( Q_{\v w, r} \right) \\
    & \subjectto             &&   \sum_{(\v w, r) \in \mathcal{A}} Q_{\v w,r} = 1 \\
    &                        &&   \sum_{(\v w, r) \in \mathcal{A}} \v w Q_{\v w,r} = \v 1 \\
    &                        &&   \sum_{(\v w, r) \in \mathcal{A}} \v t^\top \v w r Q_{\v w,r} = d,
\end{alignat*}
where $c_{\v w,r} = \sum_{i=1}^n \mathds{1}_{ \{ \v w=\v w_i, r=r_i \} }$. 

The associated Lagrange dual function is given by
\begin{align*}
    g(\gamma, \v \beta, \delta)
    &= \begin{aligned}[t]
    \inf_{\v Q} 
        \left\{
            - \sum_{(\v w, r) \in \mathcal{A}} c_{\v w, r} \log \left( Q_{\v w, r} \right) \right.
            &+ \gamma 
            \left( 
                \sum_{(\v w, r) \in \mathcal{A}} Q_{\v w,r} - 1 
            \right) \\
            &+ \v \beta^\top 
            \left( 
                \sum_{(\v w, r) \in \mathcal{A}} \v w Q_{\v w,r} - \v 1 
            \right) \\
            &+ \left.
            \delta 
            \left( 
                \sum_{(\v w, r) \in \mathcal{A}} \v t^\top \v w r Q_{\v w,r} - d 
            \right) 
        \right\}
    \end{aligned} \\
    &= \inf_{\v Q} 
    \left\{ 
        \vphantom{ \sum_{(\v w, r) \in \mathcal{A}} }
        - \left( \gamma + \v \beta^\top \v 1 + \delta d \right) \right. \\
    &\hspace{3cm}
        \left. + \sum_{(\v w, r) \in \mathcal{A}} 
        \left\{ 
            \left( 
                \gamma + \v \beta^\top \v w + \delta \v t^\top \v w r 
            \right) Q_{\v w, r} 
            - c_{\v w, r} \log \left( Q_{\v w, r} \right) 
        \right\} 
    \right\}. 
\end{align*}

To ensure that the Lagrange dual function is bounded, we must have 
\begin{align*}
    & \gamma + \v \beta^\top \v w + \delta \v t^\top \v w r \ge 0 \quad 
        \forall (\v w, r) \in \mathcal{A} \subset \mathcal{S}\\
    & \gamma + \v \beta^\top \v w + \delta \v t^\top \v w r > 0 \quad 
        \mathrm{if} \; c_{\v w,r} > 0.
\end{align*}
Note that $\mathcal{A}$ is an arbitrary subset of $\mathcal{S}$. Since we require boundedness for any such $\mathcal{A} \subset \mathcal{S}$, the constraints above must therefore hold for all $(\v w, r) \in \mathcal{S}$. Now, if $\gamma + \v \beta^\top \v w + \delta \v t^\top \v w r = 0$, then $c_{\v w,r} = 0$, and the corresponding summand involving $Q_{\v w,r}$ vanishes from the Lagrange dual function. Hence, the only relevant allocation of $Q_{\v w, r}$ with respect to the infimum is when $\gamma + \v \beta^\top \v w + \delta \v t^\top \v w r > 0$. Using standard calculus for minimization, we get
\begin{equation*}
    Q^*_{\v w,r} = \frac{c_{\v w, r}}{\gamma + \v \beta^\top \v w + \delta \v t^\top \v w r}.
\end{equation*}

Hence, the Lagrange dual function becomes
\begin{align*}
    g(\gamma, \v \beta, \v \tau)
    &= 
    - \left( \gamma + \v \beta^\top \v 1 + \delta d  \right)
    + \sum_{(\v w, r) \in \mathcal{A}} 
        \left\{ 
            c_{\v w, r} - c_{\v w, r} \log 
            \left( 
                \frac{c_{\v w, r}}{\gamma + \v \beta^\top \v w + \delta \v t^\top \v w r} 
            \right)
        \right\} \\
    &= 
    - \left( \gamma + \v \beta^\top \v 1 + \delta d \right)
    + n 
    + \sum_{(\v w, r) \in \mathcal{A}} 
        \left\{ 
            - c_{\v w, r} \log 
            \left( 
                \frac{c_{\v w, r}}{\gamma + \v \beta^\top \v w + \delta \v t^\top \v w r} 
            \right)
        \right\}. 
\end{align*}

Now consider
\begin{equation*}
    \sum_{(\v w, r) \in \mathcal{A}} 
    (\gamma + \v \beta^\top \v w + \delta \v t^\top \v w r) Q^*_{\v w,r}. 
\end{equation*}
From the $Q^*_{\v w,r}$ obtained above, this sum equals 
\begin{equation*}
    \sum_{(\v w, r) \in \mathcal{A}} c_{\v w, r} = n.
\end{equation*}
But, expanding the same sum and then applying the primal constraints yields
\begin{equation*}
    \gamma \sum_{(\v w, r) \in \mathcal{A}} Q^*_{\v w,r} 
        + \v \beta^\top \sum_{(\v w, r) \in \mathcal{A}} \v w Q^*_{\v w,r} 
        + \delta \sum_{(\v w, r) \in \mathcal{A}} \v t^\top \v w r Q^*_{\v w,r} 
    = \gamma + \v \beta^\top \v 1 + + \delta d,
\end{equation*}
where the equality holds when the dual variables are evaluated at their optimal values. Hence, $\gamma + \v \beta^\top \v 1 + \delta d = n$. This identity allows us to eliminate $\gamma$. We reparameterize the Lagrange dual function as 
\begin{equation*}
    \tilde{g}(\v \beta, \delta) = 
    g(n - \v \beta^\top \v 1 - \delta d, \; \v \beta, \; \delta), 
\end{equation*}
which simplifies to
\begin{equation*}
    \tilde{g}(\v \beta, \delta) 
    = \sum_{(\v w, r) \in \mathcal{A}} 
    c_{\v w, r} \log 
    \left(
        n + \v \beta^\top (\v w - \v 1) + \delta (\v t^\top \v w r - d)
    \right) 
    - \sum_{(\v w, r) \in \mathcal{A}} c_{\v w, r} \log(c_{\v w, r}).
\end{equation*}

Let $\v \beta \leftarrow n \v \beta$ and $\delta \leftarrow n \delta$, rescaling the dual variables to maintain numerical stability as the sample size $n$ increases. Then by strong duality, $-\log L (d)$ can be obtained by solving the Lagrange dual problem,
\begin{align*}
    - \log L (d) 
    &= \sup_{(\v \beta, \delta) \in \mathcal{D}} \tilde{g}(\v \beta, \delta) \\
    &= \sup_{(\v \beta, \delta) \in \mathcal{D}} \sum_{(\v w, r) \in \mathcal{A}} 
        c_{\v w, r} \log 
        \left(
            1 + \v \beta^\top (\v w - \v 1) + \delta (\v t^\top \v w r -  d)
        \right)
        - C \\
    &= \sup_{(\v \beta, \delta) \in \mathcal{D}} \sum_{i=1}^n 
        \log 
        \left(
            1 + \v \beta^\top (\v w_i - \v 1) + \delta (\v t^\top \v w_i r_i -  d)
        \right)
        - C,
\end{align*}
where $C = \sum_{(\v w, r) \in \mathcal{A}} c_{\v w, r} \log(c_{\v w, r}) - n \log(n)$ is an immaterial constant, and 
\begin{equation*}
    \mathcal{D} = 
    \left\{
        (\v \beta, \delta) \in \mathbb{R}^2 \times \mathbb{R}  
        \,\Big|\, 
        1 + \v \beta^\top (\v w - \v 1) + \delta (\v t^\top \v w r - d) \ge 0 
        \;\; \forall (\v w, r) \in \mathcal{S} 
    \right\}.
\end{equation*}

\section{Adaptive Sub-Support}
\label{sec:a4}

To enable efficient Bayesian computations with a fine grid, we need an adaptive sub-support for $\v v$ in which the empirical likelihood is larger than some small threshold. As in \eqref{def:sub-support} of Section~\ref{sec:sub-support}, we define the sub-support for each dimension $\mathrm{v}_j$ as follows:
\begin{equation*}
    \{\mathrm{v}_j \,|\, L(\v v) \ge c^{-1} L^* \},
\end{equation*}
where $L^*$ is the maximum empirical likelihood over the full support, and $c$ is a large positive number. Taking their Cartesian product yields the adaptive sub-support for $\v v$.

Note that this sub-support is similar to Wilks' confidence region in that both are defined by the condition that the likelihood exceeds a certain threshold. The difference lies in their construction. Wilks' confidence region applies the threshold condition jointly to the entire vector, whereas the adaptive sub-support applies this condition separately to each dimension and then takes their Cartesian product. 

Let $F^{-1}_{\chi^2_k}(q)$ denote the $q$-th quantile of the $\chi^2_k$ distribution. We choose $c$ as summarized in Table~\ref{table:choices_of_c}. The resulting adaptive sub-support corresponds to the $99.99\%$ Wilks' confidence interval or the smallest rectangle that encompasses the $99.99\%$ Wilks' confidence region. In all cases, the adaptive sub-support obtained from such choices of $c$ covers most of the relevant points in a practical sense.

\renewcommand{\arraystretch}{1.3}  
\begin{table}[ht]
\centering
\begin{tabular}{lc}
\hline\hline
Parameter           & $c$  \\ \hline \\ [-13pt]
$\v v \in [0, 1]$   & $\exp\left(\frac{1}{2} F^{-1}_{\chi^2_1}(0.9999) \right)$  \\ [5pt]
$\v v \in [0, 1]^2$ & $\exp\left(\frac{1}{2} F^{-1}_{\chi^2_2}(0.9999) \right)$  \\ [5pt]
$d \in [-1, 1]$     & $\exp\left(\frac{1}{2} F^{-1}_{\chi^2_1}(0.9999) \right)$  \\ [5pt] \hline
\end{tabular}
\caption{Choices of $c$ for the adaptive sub-support.}
\label{table:choices_of_c}
\end{table}
\renewcommand{\arraystretch}{1.0}  

Now we show how the sub-support for each dimension can be obtained by solving an optimization problem. Recall that 
\begin{equation*}
    \v v 
    = \int_{\mathcal{S}} \v w r \,\mathrm{d} \v Q(\v w, r) 
    = \sum_{(\v w, r) \in \mathcal{A}} \v w r Q_{\v w, r}, 
\end{equation*}
where $\mathcal{S} = \left(\prod_{j=1}^{\ell} [w_{j,\min}, w_{j,\max}]\right) \times [0,1]$, $\v Q$ is a probability measure on $(\mathcal{S}, \mathcal{B}(\mathcal{S}))$, $\mathcal{A} = \{ (\v w,r) \in \mathcal{S}  \,|\, \v Q(\{(\v w, r)\}) > 0 \}$, and $Q_{\v w,r} = \v Q(\{(\v w, r)\})$. 

Let $\v e_j$ be the unit vector with a one in the $j$-th component. Then the sub-support for $\mathrm{v}_j$ can be reformulated as 
\begin{equation*}
    \left\{ 
        \sum_{(\v w, r) \in \mathcal{A}} \v e_j^\top \v w r Q_{\v w, r} \Biggm|
        \sum_{(\v w, r) \in \mathcal{A}} Q_{\v w,r} = 1,
        \sum_{(\v w, r) \in \mathcal{A}} \v w Q_{\v w,r} = \v 1, 
        \sum_{(\v w, r) \in \mathcal{A}} c_{\v w, r} \log \left( Q_{\v w, r} \right) \ge \phi 
    \right\},
\end{equation*}
where $c_{\v w,r} = \sum_{i=1}^n \mathds{1}_{ \{ \v w=\v w_i, r=r_i \} }$ and $\phi = \log L^* - \log c$. 

We can then identify the sub-support for $\mathrm{v}_j$ by finding the infimum and supremum of this set with the corresponding optimization problems. We show the derivation for the infimum. The derivation for the supremum is similar and is omitted here. Consider 
\begin{alignat*}{2}
    & \minimize_{\v Q} \quad && \sum_{(\v w, r) \in \mathcal{A}} \v e_j^\top \v w r Q_{\v w, r} \\
    & \subjectto             && \sum_{(\v w, r) \in \mathcal{A}} Q_{\v w,r} = 1 \\
    &                        && \sum_{(\v w, r) \in \mathcal{A}} \v w Q_{\v w,r} = \v 1 \\
    &                        && \sum_{(\v w, r) \in \mathcal{A}} c_{\v w, r} \log \left( Q_{\v w, r} \right) \ge \phi. 
\end{alignat*}

The associated Lagrange dual function is given by
\begin{align*}
    g(\gamma, \v \beta, \kappa)
    &= \begin{aligned}[t]
    \inf_{\v Q} 
        \left\{
            \sum_{(\v w, r) \in \mathcal{A}} \v e_j^\top \v w r Q_{\v w, r} \right.
            &+ \gamma 
            \left( 
                \sum_{(\v w, r) \in \mathcal{A}} Q_{\v w,r} - 1 
            \right) \\
            &+ \v \beta^\top 
            \left( 
                \sum_{(\v w, r) \in \mathcal{A}} \v w Q_{\v w,r} - \v 1 
            \right) \\
            &+ \left.
            \kappa 
            \left( 
                \phi - \sum_{(\v w, r) \in \mathcal{A}} 
                    c_{\v w, r} \log \left( Q_{\v w, r} \right)
            \right) 
        \right\}
    \end{aligned} \\
    &= \inf_{\v Q} 
    \left\{ 
        \vphantom{ \sum_{(\v w, r) \in \mathcal{A}} }
        - \gamma - \v \beta^\top \v 1 + \kappa \phi  \right. \\
    &\hspace{3cm}
        \left. + \sum_{(\v w, r) \in \mathcal{A}} 
        \left\{ 
            \left( 
                \gamma + \v \beta^\top \v w + \v e_j^\top \v w r 
            \right) Q_{\v w,r} 
            - \kappa c_{\v w, r} \log \left( Q_{\v w, r} \right) 
        \right\} 
    \right\}. 
\end{align*}

To ensure that the Lagrange dual function is bounded, we must have 
\begin{align*}
    & \gamma + \v \beta^\top \v w + \v e_j^\top \v w r \ge 0 \quad 
        \forall (\v w, r) \in \mathcal{A} \subset \mathcal{S}\\
    & \gamma + \v \beta^\top \v w + \v e_j^\top \v w r > 0 \quad 
        \mathrm{if} \; \kappa c_{\v w,r} > 0.
\end{align*}
Note that $\mathcal{A}$ is an arbitrary subset of $\mathcal{S}$. Since we require boundedness for any such $\mathcal{A} \subset \mathcal{S}$, the constraints above must therefore hold for all $(\v w, r) \in \mathcal{S}$. Hence, the first constraint simplifies to
\begin{equation*}
    \gamma + \v \beta^\top \v w \ge 0 \quad 
        \forall (\v w, r) \in \mathcal{S}.
\end{equation*}
Now, if $\gamma + \v \beta^\top \v w + \v e_j^\top \v w r = 0$, then $\kappa c_{\v w,r} = 0$, and the corresponding summand involving $Q_{\v w,r}$ vanishes from the Lagrange dual function. Hence, the only relevant allocation of $Q_{\v w, r}$ with respect to the infimum is when $\gamma + \v \beta^\top \v w + \v e_j^\top \v w r > 0$. Using standard calculus for minimization, we get
\begin{equation*}
    Q^*_{\v w,r} = 
    \frac{\kappa c_{\v w, r}}{\gamma + \v \beta^\top \v w + \v e_j^\top \v w r}.
\end{equation*}

Hence, the Lagrange dual function becomes
\begin{align*}
    g(\gamma, \v \beta, \kappa) 
    &= - \gamma - \v \beta^\top \v 1 + \kappa \phi 
        + \sum_{(\v w, r) \in \mathcal{A}} 
        \left\{ 
        \kappa c_{\v w, r} 
        - \kappa c_{\v w, r} \log 
            \left( 
                \frac{\kappa c_{\v w, r}}{\gamma + \v \beta^\top \v w + \v e_j^\top \v w r} 
            \right) 
        \right\} \\
    &= - \gamma - \v \beta^\top \v 1 + \kappa 
        \left( 
            \phi + n 
            - \sum_{(\v w, r) \in \mathcal{A}} c_{\v w, r} \log 
            \left( 
                \frac{\kappa c_{\v w, r}}{\gamma + \v \beta^\top \v w + \v e_j^\top \v w r} 
            \right) 
        \right).
\end{align*}

Now we arrive at the Lagrange dual problem,
\begin{align*}
    & \sup_{(\gamma, \v \beta, \kappa) \in \mathcal{W}_\kappa} 
        g(\gamma, \v \beta, \kappa) \\
    =& \sup_{(\gamma, \v \beta, \kappa) \in \mathcal{W}_\kappa} 
        \left\{ 
            - \gamma - \v \beta^\top \v 1 + \kappa 
            \left(
                \phi + n 
                - \sum_{(\v w, r) \in \mathcal{A}} c_{\v w, r} \log 
                    \left( 
                        \frac
                        {\kappa c_{\v w, r}}
                        {\gamma + \v \beta^\top \v w + \v e_j^\top \v w r} 
                    \right) 
            \right) 
        \right\},
\end{align*}
where $\mathcal{W}_\kappa = \left\{ (\gamma, \v \beta, \kappa) \in \mathbb{R} \times \mathbb{R}^\ell \times \mathbb{R}^+ \,\Big|\, \gamma + \v \beta^\top \v w \ge 0  \;\; \forall (\v w, r) \in \mathcal{S} \right\}$.

Note that $\kappa > 0$. Instead of solving the Lagrange dual problem directly, we can first eliminate the dual variable $\kappa$ using standard calculus for maximization,
\begin{align*}
    \frac{\partial g}{\partial \kappa} 
    &= 
        \left(
            \phi + n 
            - \sum_{(\v w, r) \in \mathcal{A}} c_{\v w, r} \log 
            \left( 
                \frac{\kappa c_{\v w, r}}{\gamma + \v \beta^\top \v w + \v e_j^\top \v w r} 
            \right) 
        \right) 
        + \kappa 
        \left( 
            - \sum_{(\v w, r) \in \mathcal{A}} \frac{c_{\v w, r}}{\kappa} 
        \right)\\
    &= \phi - \sum_{(\v w, r) \in \mathcal{A}} c_{\v w, r} \log 
        \left( 
            \frac{\kappa c_{\v w, r}}{\gamma + \v \beta^\top \v w + \v e_j^\top \v w r} 
        \right) \\
    \frac{\partial^2 g}{\partial \kappa^2} 
    &= - \sum_{(\v w, r) \in \mathcal{A}} \frac{c_{\v w, r}}{\kappa} 
        = -\frac{n}{\kappa} 
        < 0.
\end{align*}
Hence, we have
\begin{equation*}
    \phi = \sum_{(\v w, r) \in \mathcal{A}} c_{\v w, r} \log 
        \left( 
            \frac{\kappa c_{\v w, r}}{\gamma + \v \beta^\top \v w + \v e_j^\top \v w r} 
        \right).
    \tag{$\star$} \label{id:phi}
\end{equation*}
Rearranging gives
\begin{equation*}
    \kappa = \exp 
        \left( 
            \frac
            {\phi - \sum_{(\v w, r) \in \mathcal{A}} c_{\v w, r} \log( c_{\v w, r} ) }
            {n} 
        \right) 
        \prod_{(\v w, r) \in \mathcal{A}} 
            \left( 
                \gamma + \v \beta^\top \v w + \v e_j^\top \v w r 
            \right)^{c_{\v w, r}/n}.
\end{equation*}

Now we reparameterize the Lagrange dual function as 
\begin{equation*}
    \tilde{g}(\gamma, \v \beta) = g(\gamma, \; \v \beta, \; \kappa(\gamma, \v \beta) ).
\end{equation*}
Then the Lagrange dual problem simplifies to
\begin{align*}
    & \sup_{(\gamma, \v \beta) \in \mathcal{W}} 
        \tilde{g}(\gamma, \v \beta) \\
    =& \sup_{(\gamma, \v \beta) \in \mathcal{W}} 
        \left\{ 
            - \gamma - \v \beta^\top \v 1 + \kappa(\gamma, \v \beta) \; n
        \right\} \qquad \text{(using Identity~\ref{id:phi})} \\
    =& \sup_{(\gamma, \v \beta) \in \mathcal{W}} 
        \left\{ 
            - \gamma - \v \beta^\top \v 1 
            + \exp 
            \left( 
                \frac
                {
                    \phi 
                    - \sum_{(\v w, r) \in \mathcal{A}} c_{\v w, r} \log( c_{\v w, r} ) 
                    + n \log(n) 
                }
                {n} 
            \right) 
        \right. \\
    &\hspace{6cm} 
        \left.
        \times \prod_{(\v w, r) \in \mathcal{A}} 
        \left( 
            \gamma + \v \beta^\top \v w + \v e_j^\top \v w r 
        \right)^{c_{\v w, r}/n} 
        \right\},
\end{align*}
where $\mathcal{W} = \left\{ (\gamma, \v \beta) \in \mathbb{R} \times \mathbb{R}^\ell \,\Big|\, \gamma + \v \beta^\top \v w \ge 0  \;\; \forall (\v w, r) \in \mathcal{S} \right\}$.

Finally, by strong duality, the infimum of the sub-support for $\mathrm{v}_j$ can be obtained by solving this simplified Lagrange dual problem.

\vskip 0.2in
\bibliography{CtxBandit}

\end{document}